\documentclass[runningheads,a4paper]{llncs}

\usepackage{amssymb}
\usepackage{graphicx}
\usepackage{amsmath}
\usepackage{textpos}

\usepackage{url}

\renewcommand{\u}{\mathbf{u}}
\newcommand{\x}{\mathbf{x}}
\newcommand{\xl}[1]{\x^{(#1)}}
\newcommand{\X}{\mathbf{X}}
\newcommand{\y}{\mathbf{y}}
\newcommand{\W}{\mathbf{W}}
\newcommand{\Wh}{\hat{\W}}
\newcommand{\Win}{\W_{in}}
\newcommand{\Wout}{\W_{out}}

\DeclareMathOperator*{\argmin}{arg\,min}

\begin{document}

\mainmatter  

\title{Richness of Deep Echo State Network Dynamics}

\titlerunning{Richness of Deep Echo State Network Dynamics}

\author{Claudio Gallicchio \and Alessio Micheli}

\institute{Department of Computer Science, University of Pisa,\\
Largo B. Pontecorvo, 3, 56127 Pisa, Italy\\
\path{gallicch@di.unipi.it},\path{micheli@di.unipi.it}
}

\maketitle

\begin{textblock*}{150mm}(-2cm,-6cm)
\noindent
\emph{This is a pre-print version of the following paper published in the proceedings of IWANN 2019:}\\
Gallicchio C., Micheli A. (2019) Richness of Deep Echo State Network Dynamics. In: Rojas I., Joya G., Catala A. (eds) Advances in Computational Intelligence. IWANN 2019. Lecture Notes in Computer Science, vol 11506. Springer, Cham
.
\end{textblock*}

\begin{abstract}
Reservoir Computing (RC) is a popular methodology for the efficient design of Recurrent Neural Networks (RNNs). Recently, the advantages of the RC approach have been extended to the context of multi-layered RNNs, with the introduction of the Deep Echo State Network (DeepESN) model.
In this paper, we study the quality of state dynamics in progressively higher layers of DeepESNs, using tools from the areas of information theory and numerical analysis.
Our experimental results on RC benchmark datasets reveal the fundamental role played by the strength of inter-reservoir connections to increasingly enrich the representations developed in higher layers.
Our analysis also gives interesting insights into the possibility of effective exploitation of training algorithms based on stochastic gradient descent in the RC field.

\end{abstract}

\section{Introduction}
Randomized approaches to the design of neural networks are subject of considerable attention in 
the research community nowadays 
\cite{Scardapane2017randomness,Zhang2016survey,Gallicchio2017randomized}. 
The idea of keeping the internal connections fixed
is particularly intriguing 
when considering
Recurrent Neural Networks (RNNs) \cite{Gallicchio2018randomized}. 
In this case, indeed,
the remarkable 
efficiency advantages of 
using untrained hidden weights are
coupled with the need to control the resulting system dynamics, 
to make sure 	that 
they 
can be useful for learning. 
In this context, Reservoir Computing (RC) \cite{Lukosevicius2009,Verstraeten2007} represents the reference paradigm for the randomized design of RNNs. 
A promising research line in the current development of RC is given by the exploration of its extensions to deep learning \cite{Lecun2015,Schmidhuber2015}, with the introduction of Deep Echo State Network (DeepESN) 
\cite{Gallicchio2017DeepESN} providing a refreshing perspective to the study of hierarchically structured RNNs.
On the one hand, results in \cite{Gallicchio2018design,Gallicchio2019comparison} suggested that a proper architectural design of DeepESNs can have a tremendous impact on real-world applications.
On the other hand, 
investigations on DeepESNs dynamics \cite{Gallicchio2017DeepESN,Gallicchio2017echo,Gallicchio2018local} revealed that a stacked composition of recurrent layers has the inherent ability to 
diversify the dynamical response to the driving input signals in successive levels of the architecture.
However, the effects of network setup on the quality of state dynamics in deep RC models currently remain largely unexplored.

In this paper we study the richness of state dynamics developed in successive layers of DeepESNs.
To do so, we make use of quantitative measures 
of different nature (including entropy, number of uncoupled state dynamics and condition number) and study how they vary in deeper network settings. 
Differently from the work in \cite{Gallicchio2017DeepESN}, 
here we do not consider any pre-training approach for the recurrent layers, and focus our analysis only on the intrinsic effects of  layering. 
Besides, the experimental investigation reported in this paper 
also contributes to explore
the viability of least mean squares (LMS)-based algorithms for training the output component of RC networks.

The rest of this paper is structured as follows. In Section~\ref{sec.DeepESN} we present the 
basics
of the DeepESN model, while the 
adopted 
quality measures of reservoir dynamics
are introduced in Section~\ref{sec.measures}. The experimental settings and the achieved results are reported in Section~\ref{sec.experiments}. Finally, Section~\ref{sec.conclusions} concludes the paper.

\section{Deep Echo State Networks}
\label{sec.DeepESN}
The RC approach to RNN design is based on separating the dynamical recurrent (non-linear)
component of the network, called \emph{reservoir}, from the feed-forward (linear) output layer, called \emph{readout}.
While the application of training algorithms is limited to the readout, the reservoir is initialized under stability constraints and then is left untrained, making the overall approach extremely efficient in comparison to fully trained RNN models.
The RC paradigm has several equivalent formulations,
among which the Echo State Network (ESN) \cite{Jaeger2004,Jaeger2001} is one of the most popular.
In the rest of this section we deal with deep extensions of ESNs, referring the reader interested in basic aspects of RC to the extensive overviews available in literature \cite{Lukosevicius2009,Verstraeten2007}.

A Deep Echo State Network (DeepESN) \cite{Gallicchio2017DeepESN} 
is an RC model in which the reservoir part is structured 
into a stack of layers.
The output of each reservoir layer constitutes the input for the next one in the deep architecture, and the external input signal is propagated only to the first reservoir in the stack.
A comprehensive summary of properties and recent advancements in the study of DeepESNs can be found in \cite{Gallicchio2017survey}.
\begin{figure}[bth]
  \centering
	  \includegraphics[width=1\textwidth]{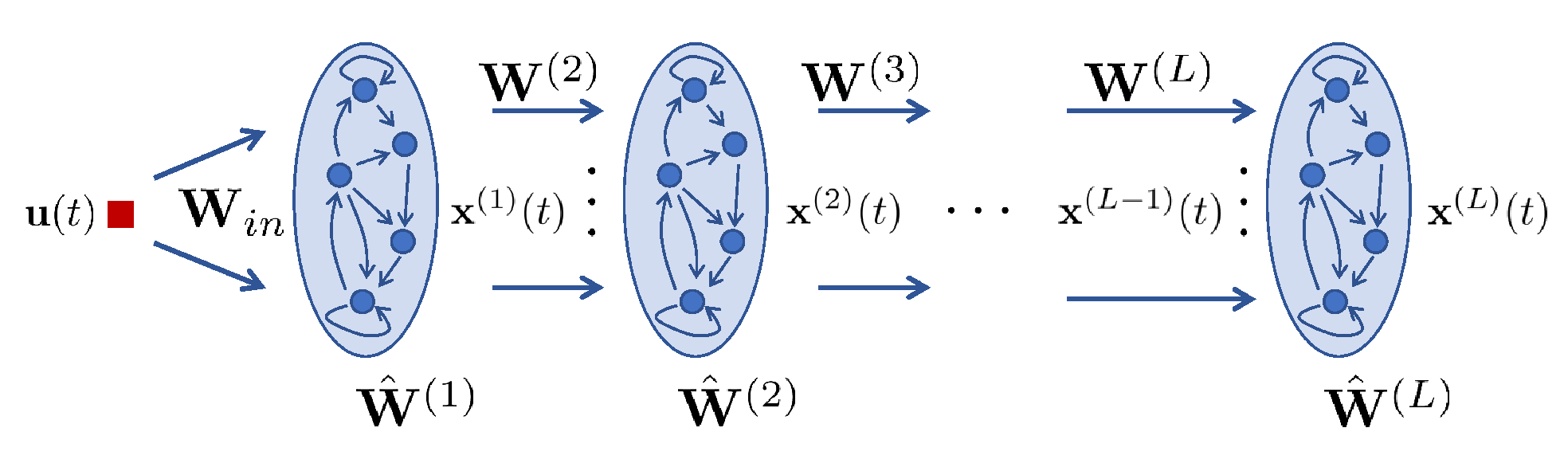}
	    \caption{The reservoir architecture of a DeepESN.}
  \label{fig.deepreservoir}
\end{figure}

In what follows, we denote the input and the output sizes respectively by $N_U$ and $N_Y$, 
the number of layers is indicated by $L$, and we make the assumption that each layer 
contains the same number of $N_R$  recurrent units.
An illustrative representation of the DeepESN reservoir architecture is given in Fig.~\ref{fig.deepreservoir}.
The state update equations of a DeepESN are described in terms of discrete-time iterated 
mappings\footnote{For the ease of notation, DeepESN equations are reported here omitting the bias terms. For a comprehensive mathematical description of DeepESN, comprising the case of leaky integrator reservoir units, the reader is referred to \cite{Gallicchio2017DeepESN}.}.
At each time-step $t$, the state of the first layer, denoted by $\xl{1}(t)$, is computed as follows:
\begin{equation}
\label{eq.layer1}
\xl{1}(t) = \tanh \big( \Win \u(t) + \Wh^{(1)} \xl{1}(t-1) \big),
\end{equation}
while for successive layers $l>1$, the state $\xl{l}(t)$ is updated according to the following equation:
\begin{equation}
\label{eq.layerl}
\xl{l}(t) = \tanh \big( \W^{(l)} \xl{l-1}(t) + \Wh^{(l)} \xl{l}(t-1) \big).
\end{equation}
In the above eqs.~\ref{eq.layer1} and \ref{eq.layerl}, 
$\Win$
is the input weight matrix, 
$\W^{(l)}$
(for $l>1$) is the inter-layer weight matrix connecting the $(l-1)$-th reservoir to the $l$-th reservoir, 
$\Wh^{(l)}$
(for $l \geq 1$) is the recurrent weight matrix for layer $l$,
and $\tanh$ denotes the element-wise application of the hyperbolic tangent non-linearity.
Typically, a zero state is used as initial condition for each layer, i.e. $\xl{l}(0) = \mathbf{0}$.
Note that whenever a single reservoir layer is considered in the architecture, i.e. if $L = 1$, the DeepESN reduces to a standard (shallow) ESN.

Following the RC paradigm,
the values in all the above mentioned reservoir weight matrices (in eqs.~\ref{eq.layer1} and \ref{eq.layerl})
are left untrained after initialization based on asymptotic stability criteria, commonly known under the name of Echo State Property (ESP) \cite{Lukosevicius2009,Yildiz2012}.
A detailed analysis of the ESP for the case of deep reservoirs is provided in one of our previous works 
\cite{Gallicchio2017echo},
to which the reader is referred for a detailed description.
Here we limit ourselves to recall that the analysis of stability of deep reservoir dynamics
essentially suggests to constrain the magnitude of the involved weight matrices, which leads to a simple initialization procedure for a DeepESN. 
Accordingly, values in $\Win$, $\{\W^{(l)}\}_{l = 2}^L$ and
$\{\Wh^{(l)}\}_{l = 1}^L$ are first chosen randomly from a uniform distribution in $[-1,1]$, and then are re-scaled to control the values of the following hyper-parameters:
\emph{input scaling} $\omega_{in} = \| \Win \|_2$, \emph{inter-layer scaling} (for $l>1$)
$\omega_{il}^{(l)} = \| \W^{(l)} \|_2$, and \emph{spectral radius}\footnote{The maximum among the eigenvalues in modulus.} (for $l\geq 1$) 
$\rho^{(l)} = \rho(\Wh^{(l)})$.
Given a driving input signal of length $T$, i.e. $\u(1), \ldots,\u(T)$, 
we find it useful
to (column-wise) collect the states developed by each reservoir layer $l$ into a state matrix:
\begin{equation}
\label{eq.X}
\X^{(l)} = \big[ \xl{l}(1) \ldots \xl{l}(T) \big].
\end{equation}

In line with the standard RC methodology,
the output of the DeepESN is computed by the readout layer as a linear combination of the reservoir activations.
As in this paper we are mainly interested in analyzing the behavior of the DeepESN locally to each layer, we study the output of the model when the readout is fed by the state developed individually by each reservoir. Accordingly, when layer $l$ is under focus, the output 
at time-step $t$, denoted by $\y^{(l)}(t)$, is computed as follows:
\begin{equation}
\label{eq.readout}
\y^{(l)}(t) = \Wout^{(l)} \xl{l}(t).
\end{equation}
In previous eq.~\ref{eq.readout}, $\Wout^{(l)}$ denotes the readout weight matrix, whose values are adjusted on a training set to solve a linear regression problem given by:
\begin{equation}
\label{eq.problem}
\| \Wout^{(l)} \X^{(l)} - \mathbf{Y}_{tg} \|_2^2,
\end{equation}
where $\X^{(l)} $ is as defined in eq.~\ref{eq.X}, and $\mathbf{Y}_{tg} = \big[ \y_{tg}(1) \ldots \y_{tg}(T) \big]$ is a matrix that collects the corresponding target signals (in a column-wise fashion).
Due to a typically large condition number of the reservoir state matrices, readout training is commonly performed by means of non-iterative methods \cite{Lukosevicius2009}.

\section{Richness of Deep Reservoir Dynamics}
\label{sec.measures}
To quantify the richness of reservoir dynamics in DeepESNs, we make use of the following measures.
\begin{description}
\item[Average State Entropy] (ASE) - From an information-theoretic perspective the richness of ESN dynamics can be effectively quantified by means of the entropy of instantaneous reservoir states \cite{Ozturk2007}. Here we employ an efficient estimator of Renyi's quadratic entropy \cite{Principe2010,Principe2000}, which, for each time step $t$ and for each layer $l$ in the deep reservoir architecture, can be computed as follows:
\begin{equation}
\label{eq.entropy}
H^{(l)}(t) = -log\big( \frac{1}{{N_R}^2}\sum_{j=1}^{N_R}  \big(\sum_{i=1}^{N_R}  \mathcal{K}(x^{(l)}_j(t) - x^{(l)}_i(t))\big)   \big),
\end{equation}
where $x^{(l)}_j(t)$ denotes the $j$-th component of $\xl{l}(t)$, and $\mathcal{K}$ is a gaussian kernel (whose size is obtained by shrinking the standard deviation of instantaneous reservoir activations by a factor of $0.3$, in analogy to \cite{Ozturk2007}).
Given an input signal of length $T$, we compute the average state entropy (ASE) of layer $l$ as the time-averaged value of the Renyi's quadratic entropy in eq.~\ref{eq.entropy}:
\begin{equation}
\label{eq.ase}
ASE^{(l)} = \frac{1}{T}\sum_{t = 1}^T H^{(l)}(t) .
\end{equation}
\item[Uncoupled Dynamics] (UD) - 
It is a known fact in RC literature that reservoir units exhibit behaviors that are strongly coupled among each other \cite{Lukosevicius2009,Xue2007}. In one of our previous works \cite{Gallicchio2010} we experimentally showed that this phenomenon can be understood in terms of the inherent Markovian characterization of ESN dynamics \cite{Gallicchio2011NN,Tino2007}. Essentially, the stronger is the contractive characterization of a reservoir (i.e. the smaller is the Lipschitz constant of its state transition function) the stronger is the observed redundancy of its units activations, and the poorer is the quality of reservoir dynamics provided to the readout learner. 
Following this spirit, we propose to evaluate the richness of reservoirs by measuring the actual dimensionality of (linearly) uncoupled state dynamics. To this aim, here we take a simple approach consisting in computing the number of the principal components (i.e. orthogonal directions of variability) that are able to explain the most of the variance in the reservoir state space. 
Specifically, given an input signal of length $T$, the $l$-th reservoir layer is driven into a set of states collected into the state matrix $\X^{(l)}$ (see eq. \ref{eq.X}). Denoting by  $\sigma_1^{(l)}, \ldots, \sigma_{N_R}^{(l)}$ the singular values of $\X^{(l)}$ in decreasing order (i.e., the eigenvalues of the covariance matrix of reservoir units activations), the (normalized) relevance of the $i$-th principal component can be computed as follows:
\begin{equation}
\label{eq.relevance}
R_i^{(l)} = \frac{\sigma_i^{(l)}}{\sum_{j = 1}^{N_R} \sigma_j^{(l)}}.
\end{equation}
Based on this, the uncoupled dynamics (UD) indicator of the $l$-th reservoir is given by:
\begin{equation}
\label{eq.dimensionality}
UD^{(l)} = \argmin\limits_{d}  \Big\lbrace 
\sum_{k=1}^d R_k^{(l)}  \; | \;  \sum_{k=1}^d R_k^{(l)} \geq \mathcal{A}
\Big\rbrace,
\end{equation}
where $\mathcal{A}$ ranges in $(0,1]$ and expresses the desired amount of explained variability. In this paper we considered $\mathcal{A} = 0.9$, meaning that ${UD}^{(l)}$ is the number of linearly uncoupled directions that explain the $90\%$ of the state space variability.

\item[Condition Number] ($\kappa$) - Another well-known measure for reservoir quality is given by the conditioning of the resulting learning problem for the readout learner (see eq.~\ref{eq.problem}). Conventional ESNs are known to suffer from poor conditioning \cite{Lukosevicius2009,Jaeger2005Riddles} (i.e., high eigenvalue spread), which (among the other downsides) prevents the use of  efficient LMS-based learning algorithms employing stochastic gradient descent \cite{Haykin2009} in RC contexts.
In this paper, we study the conditioning of the state representation developed in successive levels of DeepESNs. To this end, for each layer $l$ in the deep reservoir architecture, we compute the condition number of its reservoir state matrix $\X^{(l)}$, as follows:
\begin{equation}
\label{eq.condition}
\kappa^{(l)} = \frac{\sigma_1^{(l)}}{\sigma_{N_R}^{(l)}},
\end{equation}
where $\sigma_1^{(l)}$ and $\sigma_{N_R}^{(l)}$ are respectively the largest and the smallest singular values of $\X^{(l)}$, with smaller values of $\kappa^{(l)}$ indicating richer reservoirs.
\end{description}

\section{Experiments}
\label{sec.experiments}
In this section we report the outcomes of our experimental analysis. Specifically, details on the considered datasets and experimental settings are given in Section~\ref{sec.settings}, whereas numerical results are described in Section~\ref{sec.results}.

\subsection{Datasets and Experimental Settings}
\label{sec.settings}
We considered two well-known benchmark datasets in the RC area, both involving univariate time-series (i.e., $N_U = N_Y = 1$).
Although the major focus of our analysis is on the evaluation of deep reservoir dynamics excited by the input (irrespective of the target output values), 
the 
datasets are also taken in into account 
for the definition of regression tasks.

The first dataset is related to the prediction of a 10th order non-linear auto-regressive moving average (NARMA) system, where at each time-step the input $u(t)$ is randomly sampled from a uniform distribution in $[0,0.5]$, and the target $y_{tg}(t)$ is given by the following equation:
\begin{equation}
\label{eq.narma10}
y_{tg}(t) = 0.3 \, y_{tg}(t-1) + 0.05 \,y_{tg}(t-1) \, \big(\sum_{i = 1}^{10}y_{tg}(t-i)\big) + 1.5 \, u(t-10) \, u(t-1) + 0.1.
\end{equation}

The second dataset is the Santa Fe Laser dataset \cite{Weigend2018time}, consisting in a time-series of sampled intensities from a far-infrared laser in chaotic regime. The dataset enables the definition of a next-step prediction task where $y_{tg}(t) = u(t+1)$ for each time-step $t$. 
As a minimal pre-processing step, the original values present in the Laser dataset were scaled by a factor of $0.01$.

For both the NARMA and the Laser datasets, we considered sequences of length $T = 5000$ to assess the richness of reservoir dynamics. The same data was used as training set in the 
regression
experiments, where the continuation of the respective temporal sequences was considered as test set (of length $5000$ for NARMA, and of length $5092$ for Laser).
In all the experiments the first $1000$ time-steps were used as transient to washout the initial conditions from state dynamics. 

In our experiments, we considered DeepESNs with a number of layers $L$ ranging from $1$ to $5$.
All reservoir layers contained $N_R = 100$ fully connected recurrent units, and shared the same values of 
the spectral radius $\rho$ and inter-layer scaling $\omega_{il}$ (i.e., $\rho = \rho^{(1)} = \ldots = \rho^{(L)}$, $\omega_{il} = \omega_{il}^{(2)} = \ldots = \omega_{il}^{(L)}$).
For every DeepESN hyper-parameterization, results were averaged (and standard deviations computed) over 
$15$ networks realizations (with the same values of the hyper-parameters, but random initialization).

We focused our experimental analysis on the behavior of reservoir dynamics in progressively higher layers. 
As such, all the measures of state richness detailed in Section~\ref{sec.measures} (as well as the predictive performance) were computed on a layer-wise basis, i.e. individually for each layer in the deep architecture.

\subsection{Results}
\label{sec.results}

\begin{figure}[p]
  \centering
	  \includegraphics[width=0.495\textwidth]{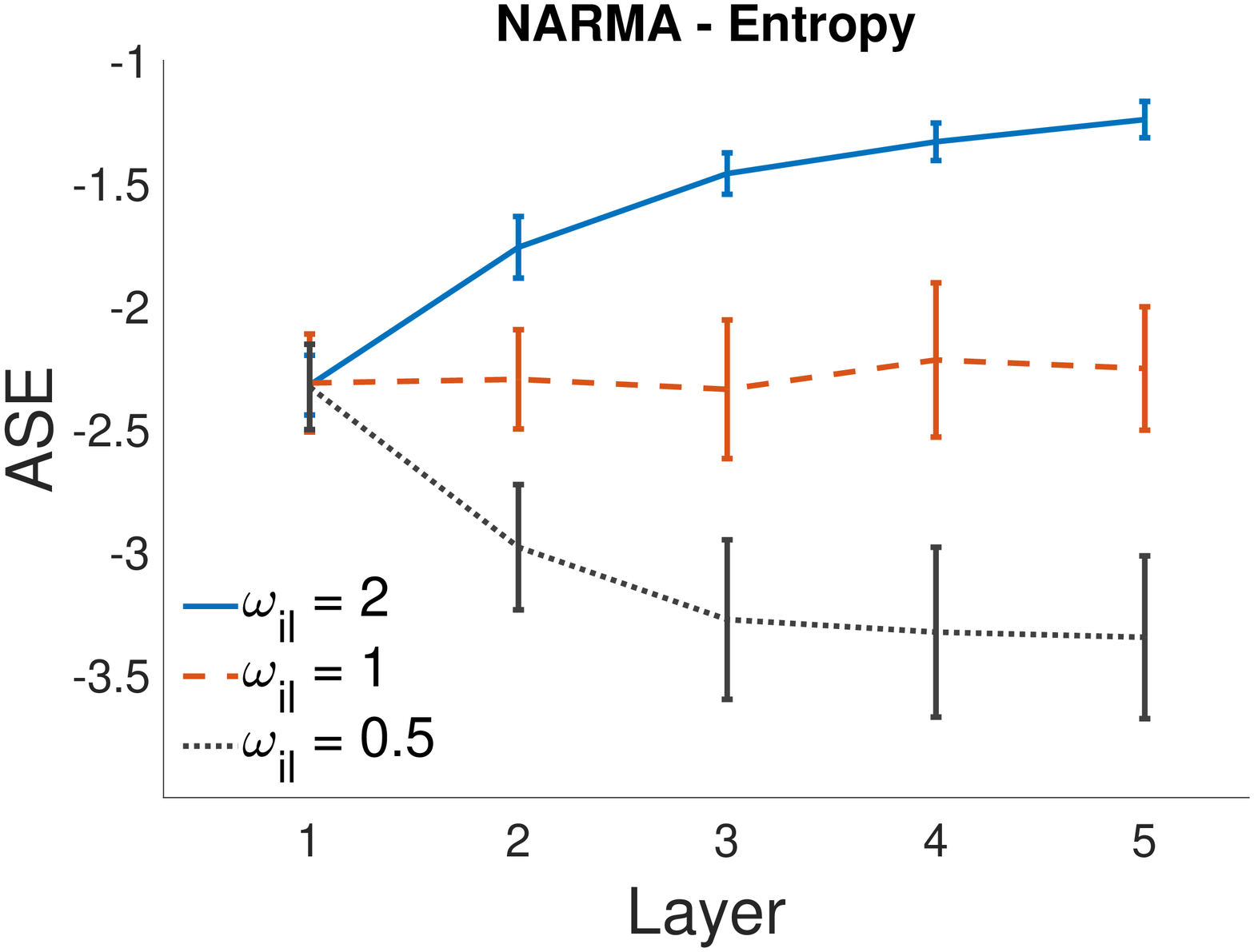}
	  \includegraphics[width=0.495\textwidth]{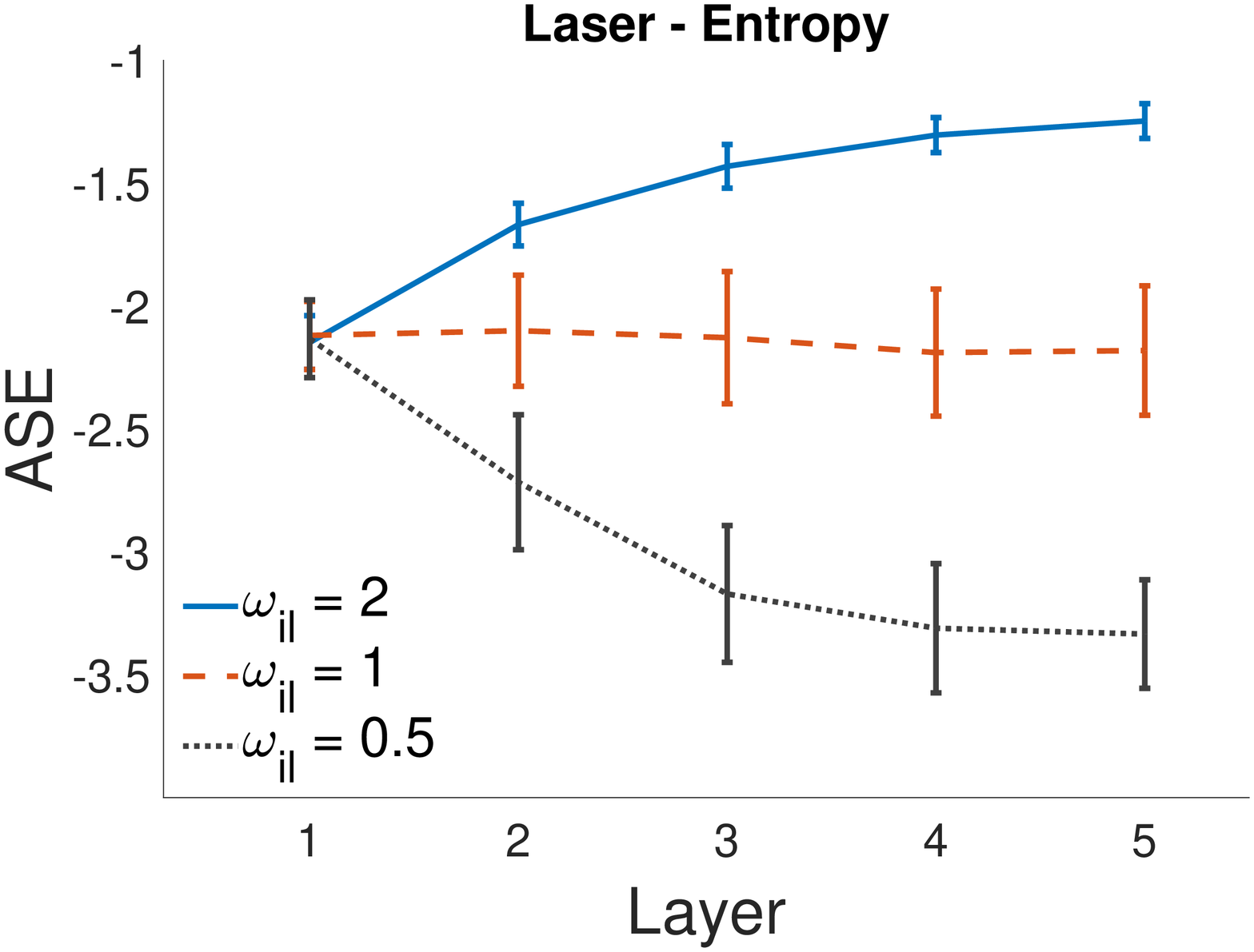}
      \\
      \vspace{0.5cm}
   	  \includegraphics[width=0.495\textwidth]{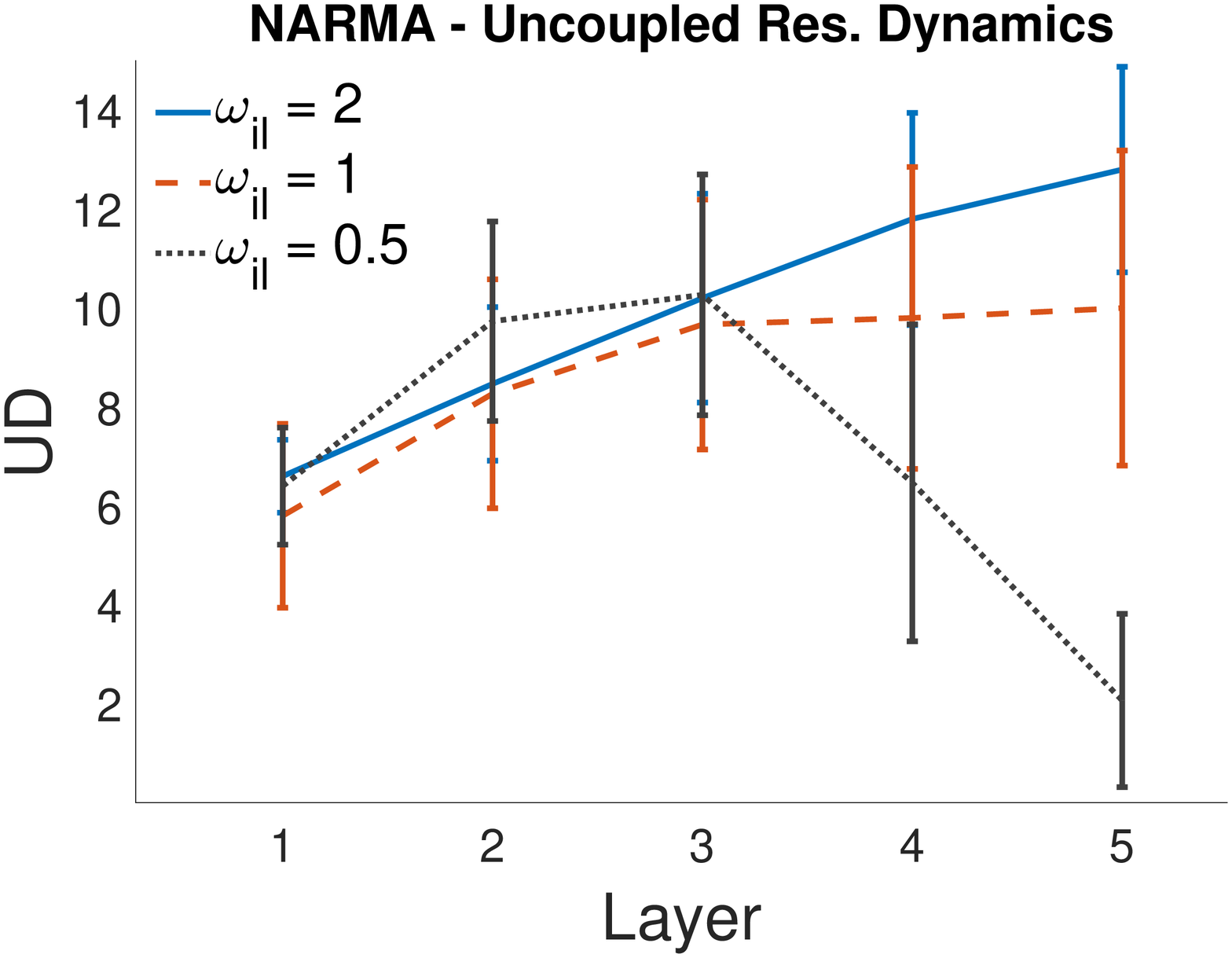}
	  \includegraphics[width=0.495\textwidth]{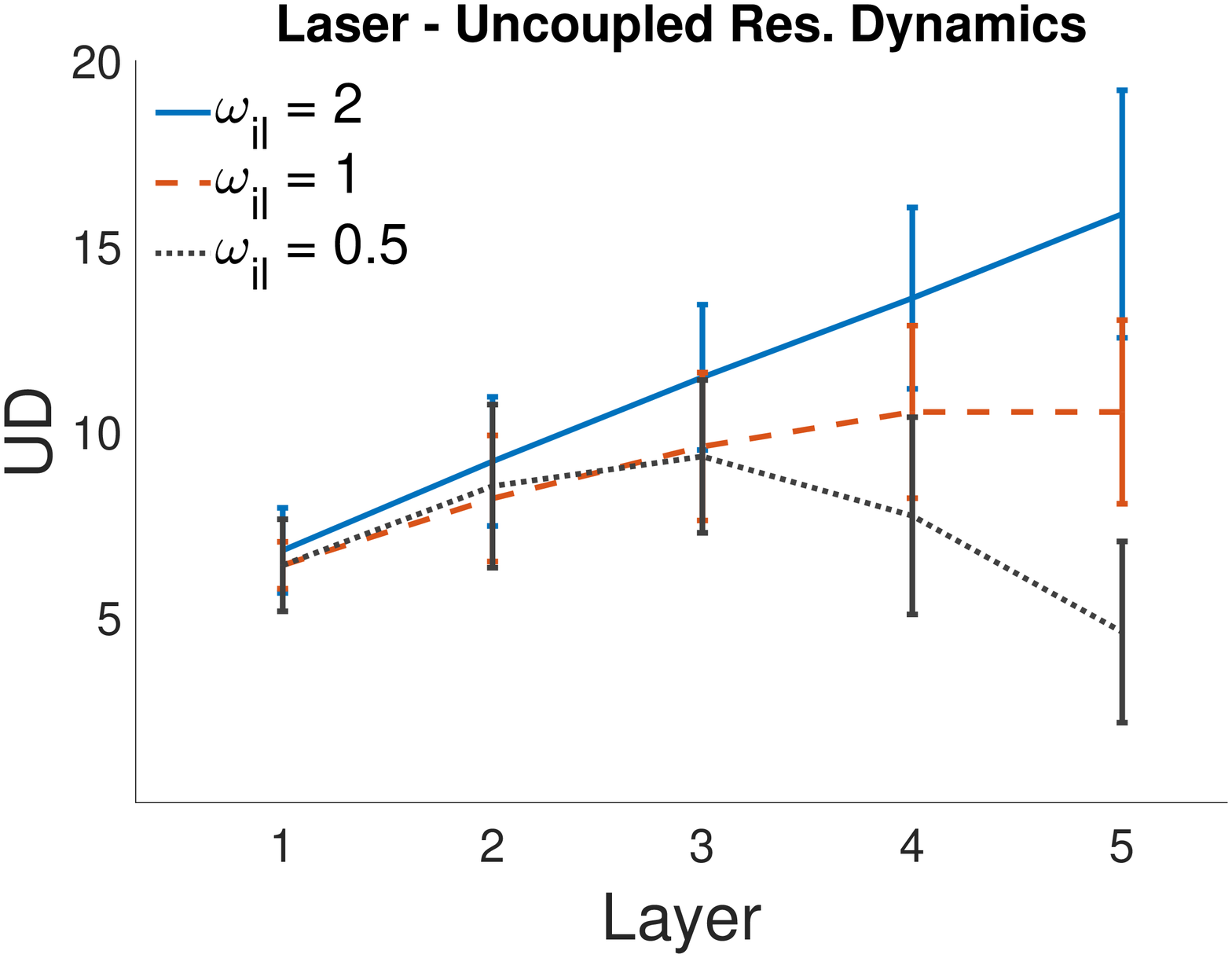}
	  \\
      \vspace{0.5cm}
   	  \includegraphics[width=0.495\textwidth]{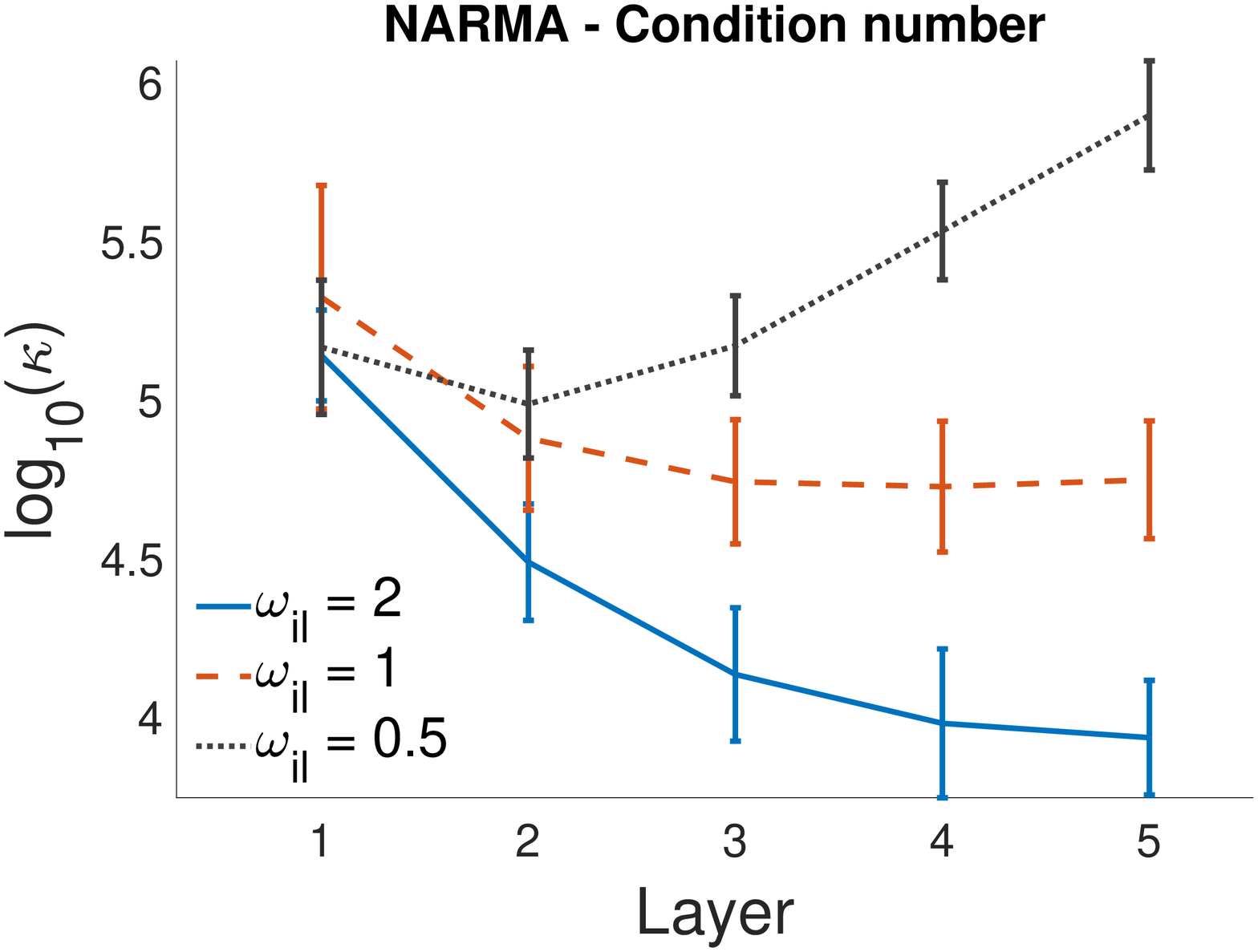}
	  \includegraphics[width=0.495\textwidth]{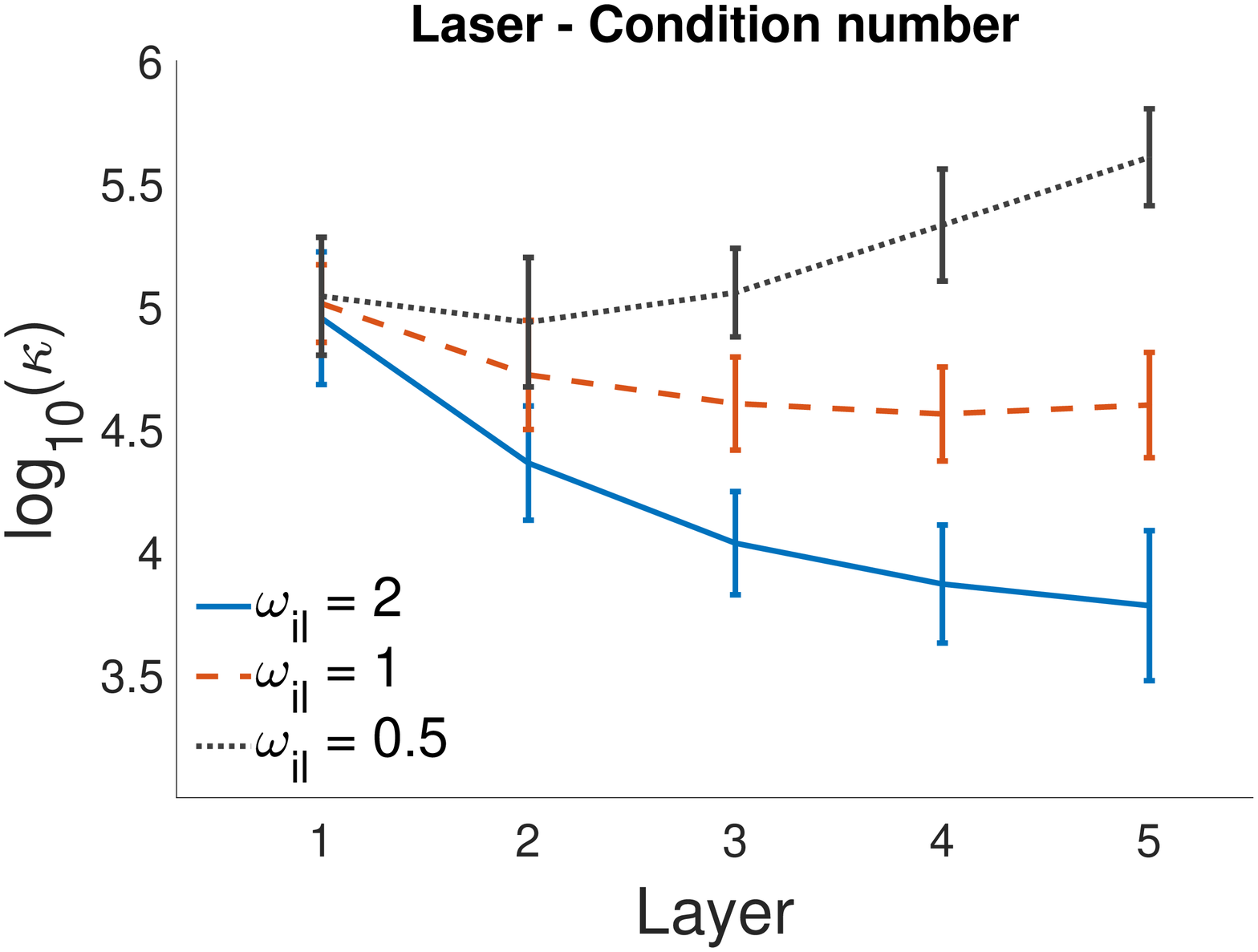}

  \caption{Measures of reservoir richness in successive reservoir layers of DeepESN.
           Results are achieved using reservoir layers with $100$ fully connected units, $\rho = 0.9$, $\omega_{in} = 1$, 
           and varying $\omega_{il}$ in $\{0.5, 1, 2\}$. 
           Standard deviations correspond to different reservoir realizations with the same hyper-parameters.
		   \emph{First row}: Average State Entropy (ASE, the higher the better).
		   \emph{Second row}: Uncoupled Reservoir Dynamics (UD, the higher the better).
		   \emph{Third row}: Condition number in log scale ($\log_{10}(\kappa)$, the lower the better).
           }
  \label{fig.richness}
\end{figure}

Fig.~\ref{fig.richness} shows 
the quality measures of DeepESN dynamics, computed 
in correspondence of progressively higher layers of the architecture.
In particular, results correspond to DeepESNs with reservoir layers hyper-parameterized by 
input scaling $\omega_{in} = 1$ and
spectral radius $\rho = 0.9$ (a value that is of common use in ESN practice).
Results are shown for values of $\omega_{il} = $ 2, 1, and 0.5, as representatives for the cases of strong, medium and weak inter-layer connectivity strength, respectively.

Results indicate that when strong inter-layer connections are used,
the quality of state representations in DeepESNs improves significantly in progressively higher layers. In this case 
($\omega_{il} = 2$ in our experiments) we can appreciate a progressive increase of both the state entropy ($ASE$, first row in Fig.~\ref{fig.richness}) and the number of relevant directions of state variability ($UD$, second row in Fig.~\ref{fig.richness}), corresponding to a substantial decrease of the condition number ($\log_{10}(\kappa)$, third row in Fig.~\ref{fig.richness}). 
With the increasing layer number we can also observe a saturation trend in the improvement of both entropy and condition number.
Interestingly, from Fig.~\ref{fig.richness} we can note that the marked enrichment of reservoir dynamics is generally observed only for strong connections between consecutive layers.
For medium  values of $\omega_{il}$ we see that the reservoir quality improves only slightly or remains substantially unchanged ($\omega_{il} = 1$ in our experiments), and for small values of $\omega_{il}$ it eventually gets worse ($\omega_{il} = 0.5$ in our experiments).
Overall, our results point out the major role played by the inter-layer scaling parameter $\omega_{il}$ in determining the trend of improvement/worsening of reservoir quality in deeper network settings. In this sense, the impact of other RC parameters, such as the spectral radius $\rho$ and the input scaling $\omega_{in}$, resulted to be much less relevant and is not shown here for brevity\footnote{Changing the value of $\rho$ resulted in  globally scaling (up/down for larger/smaller values, respectively) the achieved results for every layer. Changing the value of $\omega_{in}$ affected only the 
results in the first  layer (scaling it up/down for larger/smaller values, respectively).
In any case, changes in the values of $\rho$ and $\omega_{in}$ did not affect the quality of  the 
results 
in Fig.~\ref{fig.richness} at the increase of network's depth.}.

\begin{figure}[btp]
  \centering
   	  \includegraphics[width=0.495\textwidth]{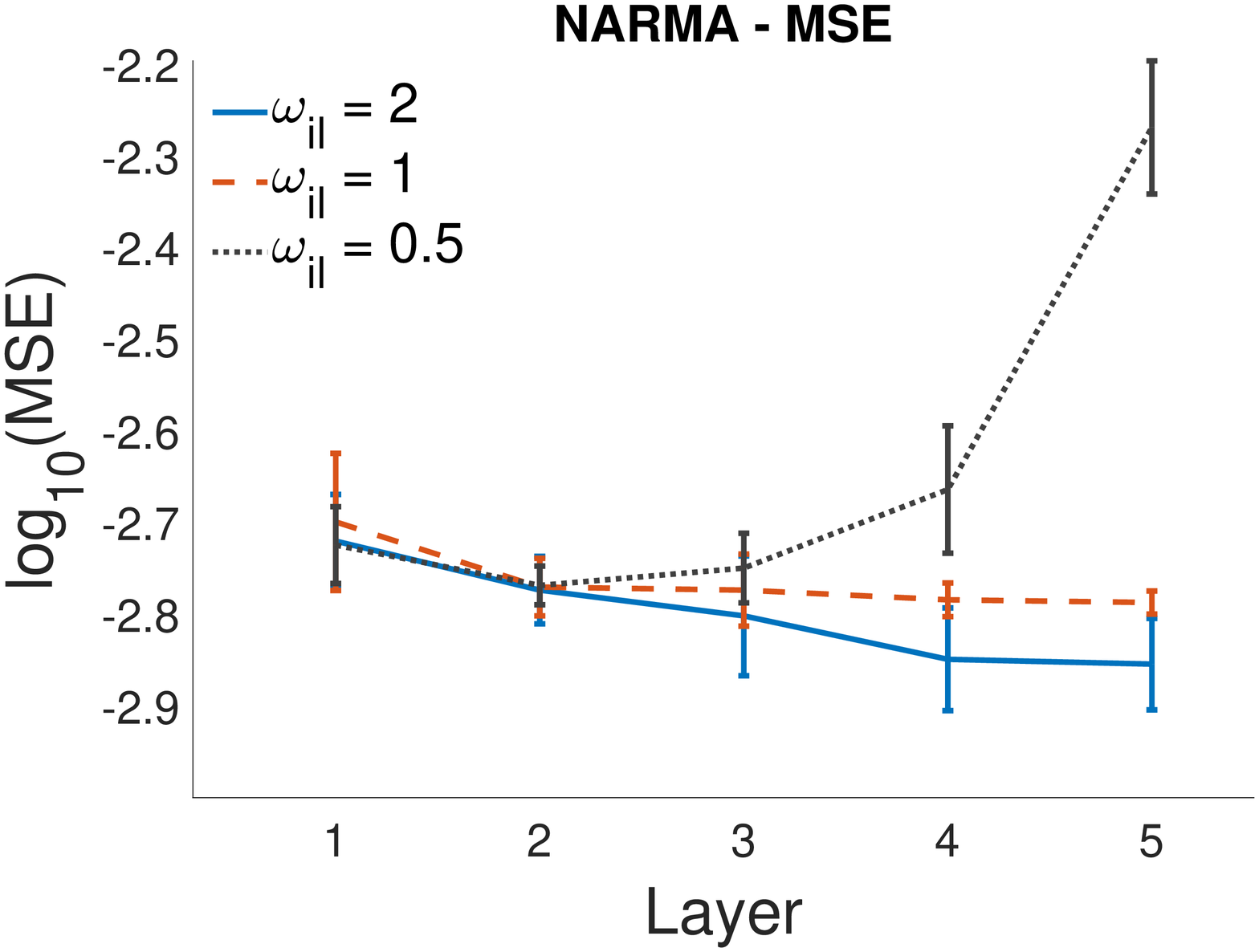}
	  \includegraphics[width=0.495\textwidth]{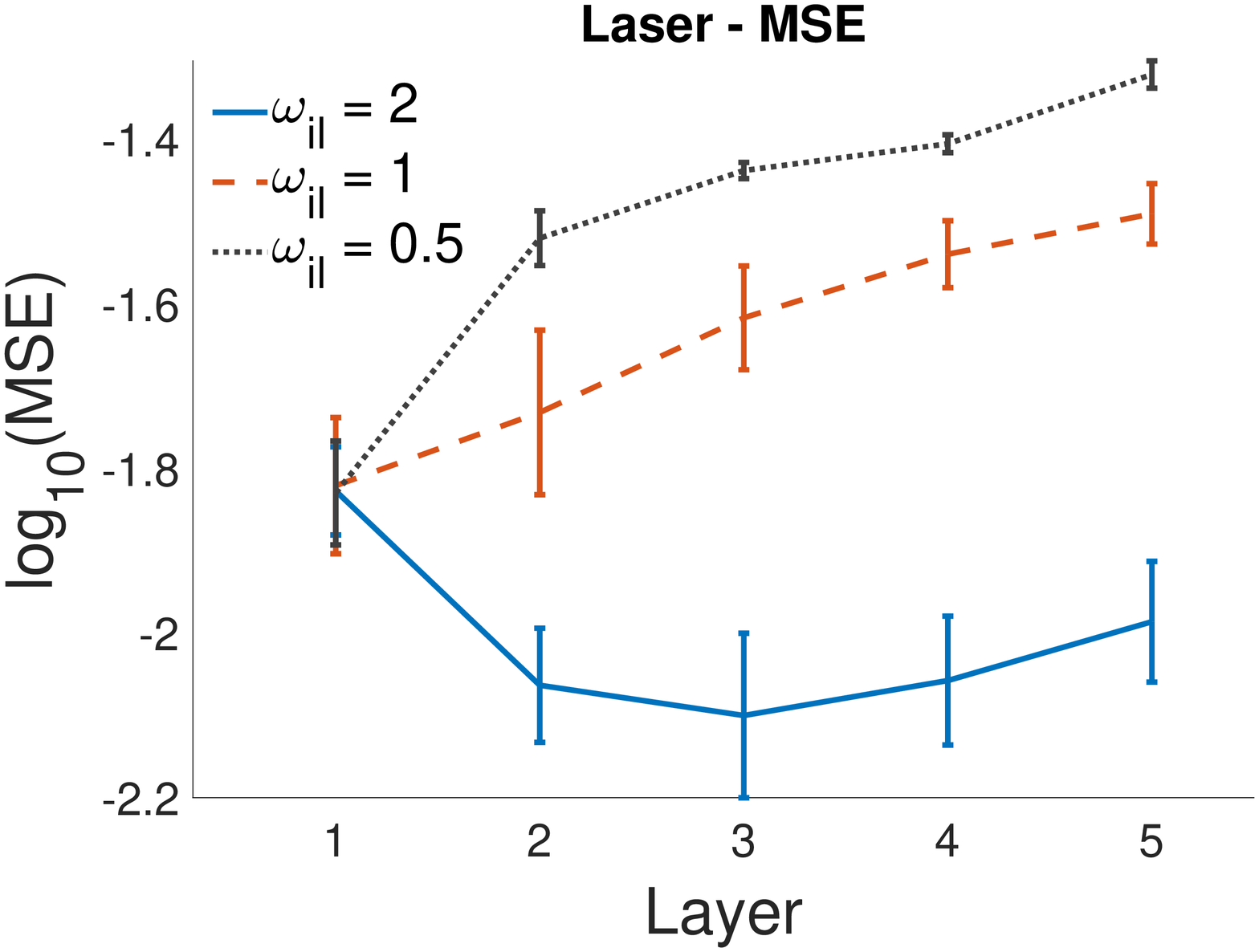}

  \caption{MSE (in $\log_{10}$ scale) on the test set for increasing DeepESN depth (the smaller the better). Results are achieved by training the readout individually on each layer, using a simple LMS algorithm. Reservoir hyper-parameters are the same as in Fig.~\ref{fig.richness}. Standard deviations correspond to different reservoir realizations with the same hyper-parameters. Test set results reported here reflect the same trend of training set results (not shown for the sake of conciseness).}
  \label{fig.mse}
\end{figure}

From the perspective of RC training algorithms, an interesting consequence of the possible decrease of the condition number in higher reservoir layers (third row in Fig.~\ref{fig.richness}) 
is that it makes potentially more suitable the application of computationally cheap stochastic gradient descent algorithms. 
To start exploring this possibility, we performed a further set of experiments, training the readout component fed by the reservoir states at individual layers, applying a basic LMS algorithm with learning rate $\eta = 0.01$ for $5000$ epochs. 
Note that our aim was not to optimize the training algorithm to achieve the highest possible performance on the specific tasks, rather we focused on analyzing the LMS performance when progressively higher reservoir layers are considered (from the qualitative view-point).
Moreover, notice that when considering progressively more layers the cost of readout training remains constant (as the input for the readout has the same dimensionality in all the cases), and the only extra-cost is that of the involved state computation in the lower reservoir layers. 
Under this perspective, the operation of the lower layers in the deep recurrent architecture can be intended as a composition of filters that progressively pre-process the data for the final reservoir level.
The MSE achieved on the test sets of the two considered datasets is shown in Fig.~\ref{fig.mse} (under the same settings considered in Fig.~\ref{fig.richness}).

Results in Fig.~\ref{fig.mse} nicely agree with those in Fig.~\ref{fig.richness}, further indicating the relevant role played by the inter-layer scaling also in terms of LMS effectiveness
at the increase of network's depth.
In particular, for medium/low inter-layer strength ($\omega_{il} = 1, 0.5$) we see that the error stays essentially constant or even severely increase, while stronger connections among layers ($\omega_{il} = 2$) result in a progressive error drop.
As a side observation, on the Laser dataset we can notice that the highest performance is achieved in correspondence of the $3$-rd reservoir layer, after which the performance starts degrading slightly 
(a trend confirmed also on the training set, not reported in Fig.~\ref{fig.mse} for the sake of conciseness). This behavior reflects the fact that the performance on a supervised learning task clearly depends on the task characterizations in a broad sense (including the target properties), and it is in line with the observations already made in \cite{Boedecker2012information,Gallicchio2018local} in relation to the memorization skills of reservoirs.

\section{Conclusions}
\label{sec.conclusions}
In this paper we have studied the richness of reservoir representations in 
DeepESN architectures.
Our empirical analysis, conducted on benchmark datasets in the RC area, pointed out the key role played by the strength of connections between the successive layers of recurrent units. Our major finding is that a strong inter-layer connectivity is required in order to determine a progressive enrichment of the state dynamics as the network's depth is increased. This outcome is interestingly related to recent results in the context of reservoirs composed of spiking neurons \cite{Zajzon2018transferring}, showing the importance of inter-reservoirs connections to properly propagate information across the levels of a deep recurrent architecture. 

Our empirical results indicate that in presence of strong inter-layer connections, reservoirs in higher layers (i.e., further away from the input) are able to develop internal representations featured by increasing entropy and with higher intrinsic (linearly uncoupled) dimensionality, at the same time leading to a decrease in the condition number that characterizes the resulting state matrices.
Interestingly, this latter observation is paired with an improved effectiveness of LMS 
for training the readout, which mitigates the widely known issue with the applicability of gradient descent algorithms in the context of RC.

While already revealing on the potentialities of deep architectures in the RC context, the work presented in this paper allows us to pave the way for promising future research developments. In this regard, we find particularly interesting to extend the application of the analysis tools delineated here
to drive the architectural design of DeepESNs in challenging real-world tasks. A starting point in this sense might be represented by a fruitful exploitation of the saturation effects shown for the quality of reservoirs in deeper networks settings. Besides, the insights on the improved amenability of LMS in DeepESNs give a further stimulus to investigate the use of efficient state-of-the-art training algorithms based on stochastic gradient descent in the RC field.

\bibliographystyle{splncs03}
\bibliography{references}

\end{document}